\documentclass[11pt,a4paper]{article}
\usepackage[hyperref]{naaclhlt2019}
\usepackage{times}
\usepackage{latexsym}
\usepackage{url}

\aclfinalcopy 
\def\aclpaperid{1467} 

\usepackage{booktabs}
\usepackage{comment}

\usepackage{subcaption}
\usepackage{array}
\usepackage{amsmath}

\newcommand{\iqadataset}{\mbox{IQUAD V1}}
\newcommand{\thornav}{\mbox{THOR-Nav}}

\DeclareMathAlphabet\mathbfcal{OMS}{cmsy}{b}{n}
\newcommand{\model}{$\mathcal{M}$}
\newcommand{\mmodel}{\mathcal{M}}
\newcommand{\visionin}{$\mathcal{V}$}
\newcommand{\mvisionin}{\mathcal{V}}
\newcommand{\languagein}{$\mathcal{L}$}
\newcommand{\mlanguagein}{\mathcal{L}}
\newcommand{\historyin}{$a$}
\newcommand{\mhistoryin}{a}
\newcommand{\mzero}{\textcolor{red}{\vec{0}}}

\newcommand{\navL}{$\mathcal{A}+\mathcal{L}$}  
\newcommand{\navV}{$\mathcal{A}+\mathcal{V}$}  
\newcommand{\navA}{$\mathcal{A}$}  
\newcommand{\qaL}{$\mathcal{L}$ \textsc{only}}  
\newcommand{\qaV}{$\mathcal{V}$ \textsc{only}}  
\newcommand{\qaA}{$\mathcal{A}$ \textsc{only}}  

\newcommand{\bad}[1]{\textcolor{blue}{\textbf{#1}}}

\usepackage{graphicx}  
\usepackage{multirow}  

\begin{document}

\title{
Shifting the Baseline:\\Single Modality Performance on Visual Navigation \& QA
}
\author{Jesse Thomason\ \ \ \ \ \ \ \ \ \  Daniel Gordon\ \ \ \ \ \ \ \ \ \  Yonatan Bisk\\
Paul G. Allen School of Computer Science and Engineering\\
\url{jdtho@cs.washington.edu}}
\maketitle

\begin{abstract}
We demonstrate the surprising strength of unimodal baselines in multimodal domains, and make concrete recommendations for best practices in future research.
Where existing work often compares against \textit{random} or \textit{majority class} baselines, 
we argue that unimodal approaches better capture and reflect dataset biases and therefore
provide an important comparison when assessing the performance of multimodal techniques.
We present unimodal ablations on three recent datasets in visual navigation and QA, seeing an up to 29\% absolute gain in performance over published baselines.
\end{abstract}

\section{Introduction}
All datasets have biases.
Baselines should capture these regularities so that outperforming them indicates a model is actually solving a task.
In multimodal domains, bias can occur in any subset of the modalities.
To address this, we argue it is not sufficient for researchers to provide random or majority class baselines; instead we recommend presenting results for unimodal models.
We investigate visual navigation and question answering tasks, where agents move through simulated environments using egocentric (first person) vision.  
We find that unimodal ablations (e.g., language only) in these seemingly multimodal tasks can outperform corresponding full models (\S \ref{ssec:ex_nav}).

This work extends observations made in both the Computer Vision~\cite{goyal18,cirik:naacl18} and Natural Language~\cite{mudrakarta:acl18, glockner-shwartz-goldberg:2018:Short, hypothesis-only-baselines-in-natural-language-inference, N18-2017, D18-1546} communities that complex models often perform well by fitting to simple, unintended correlations in the data, bypassing the complex grounding and reasoning that experimenters hoped was necessary for their tasks.

\begin{figure}[t]
\centering
\includegraphics[width=\columnwidth]{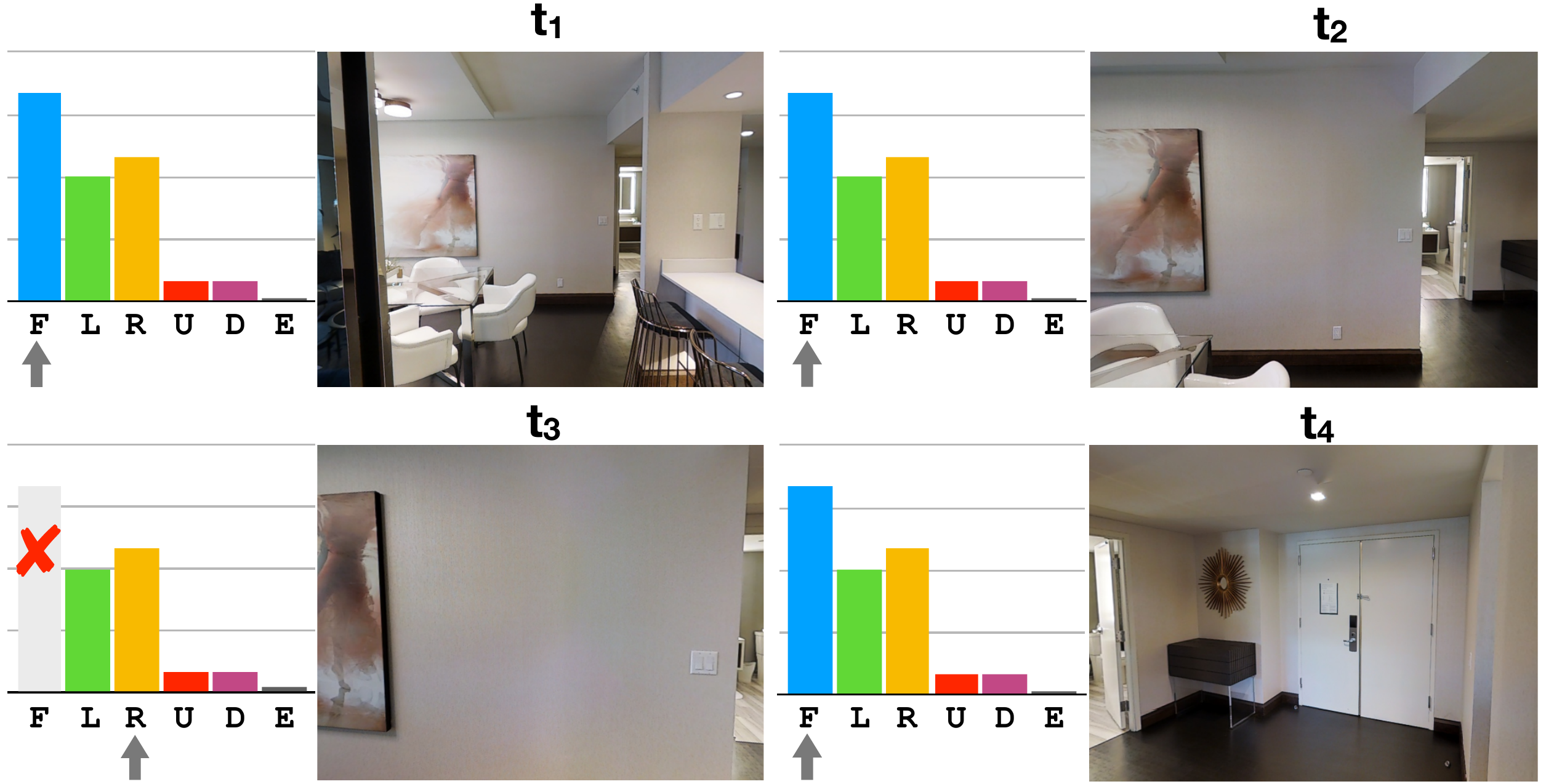}\\
\vspace{-5pt}\begin{small}Actions: \textbf{\texttt{F}}orward, turn \textbf{\texttt{L}}eft \& \textbf{\texttt{R}}ight, tilt \textbf{\texttt{U}}p \& \textbf{\texttt{D}}own, \textbf{\texttt{E}}nd\end{small}
\caption{
Navigating without vision leads to sensible navigation trajectories in response to commands like ``walk past the bar and turn right''.
At $t_3$, ``forward'' is unavailable as the agent would collide with the wall.
}
\label{fig:matterport_hallway}
\end{figure}

We ablate models from three recent papers:
(1) navigation (Figure~\ref{fig:matterport_hallway}) using images of real homes paired with crowdsourced language descriptions~\cite{anderson:cvpr18}; and (2, 3) navigation and egocentric question answering~\cite{gordon:cvpr18,das:eqa17} in simulation with synthetic questions.
We find that unimodal ablations often outperform the baselines that accompany these tasks.

\paragraph{Recommendation for Best Practices:}
Our findings show that in the new space of visual navigation and egocentric QA, all modalities, even an agent's action history, are strongly informative.  Therefore, while many papers ablate either language \textit{or} vision, new results should ablate \textit{both}.
Such baselines expose possible gains from unimodal biases in multimodal datasets irrespective of training and architecture details.

\section{Ablation Evaluation Framework}

In the visual navigation and egocentric question answering tasks, at each timestep an agent receives an observation and produces an action.
Actions can move the agent to a new location or heading (e.g., {\it turn left}), or answer questions (e.g., {\it answer `brown'}).
At timestep $t$, a multimodal model \model{} takes in a visual input \visionin$_t$ and language question or navigation command \languagein{} to predict the next action $a_t$.
The navigation models we examine also take in their action from the previous timestep, \historyin{}$_{t-1}$, and `minimally sensed' world information $W$ specifying which actions are available (e.g., that {\it forward} is unavailable if the agent is facing a wall).
\begin{equation}
    a_t \leftarrow \mmodel(\mvisionin_t, \mlanguagein, \mhistoryin_{t-1}; W)
\end{equation} 

In each benchmark, \model{} corresponds to the author's released code and training paradigm.
In addition to their full model, we evaluate the role of each input modality by removing those inputs and replacing them with zero vectors.
Formally, we define the full model and three ablations:

\vspace{-10pt}
\begin{align}
    &\text{Full Model} & \mathrm{is} && \mmodel(\mvisionin_t, \mlanguagein, \mhistoryin_{t-1}; W) \\
    &\mathcal{A} & \mathrm{is} && \mmodel(\hspace{3pt}\mzero\hspace{2pt},\hspace{1pt} \mzero\hspace{2pt}, \mhistoryin_{t-1}; W) \\
    &\mathcal{A+V} & \mathrm{is}& & \mmodel(\mvisionin_t,\hspace{1pt} \mzero\hspace{2pt}, \mhistoryin_{t-1}; W)\\
    &\mathcal{A+L} & \mathrm{is}& & \mmodel(\hspace{3pt}\mzero\hspace{2pt}, \mlanguagein, \mhistoryin_{t-1}; W)
\end{align}
corresponding to models with access to $\mathbfcal{A}$ction inputs, $\mathbfcal{V}$ision inputs, and $\mathbfcal{L}$anguage inputs.
These ablations preserve the architecture and number of parameters of \model{} by changing only its inputs.

\section{Benchmark Tasks}
\begin{figure}[t]
\includegraphics[width=0.9\linewidth]{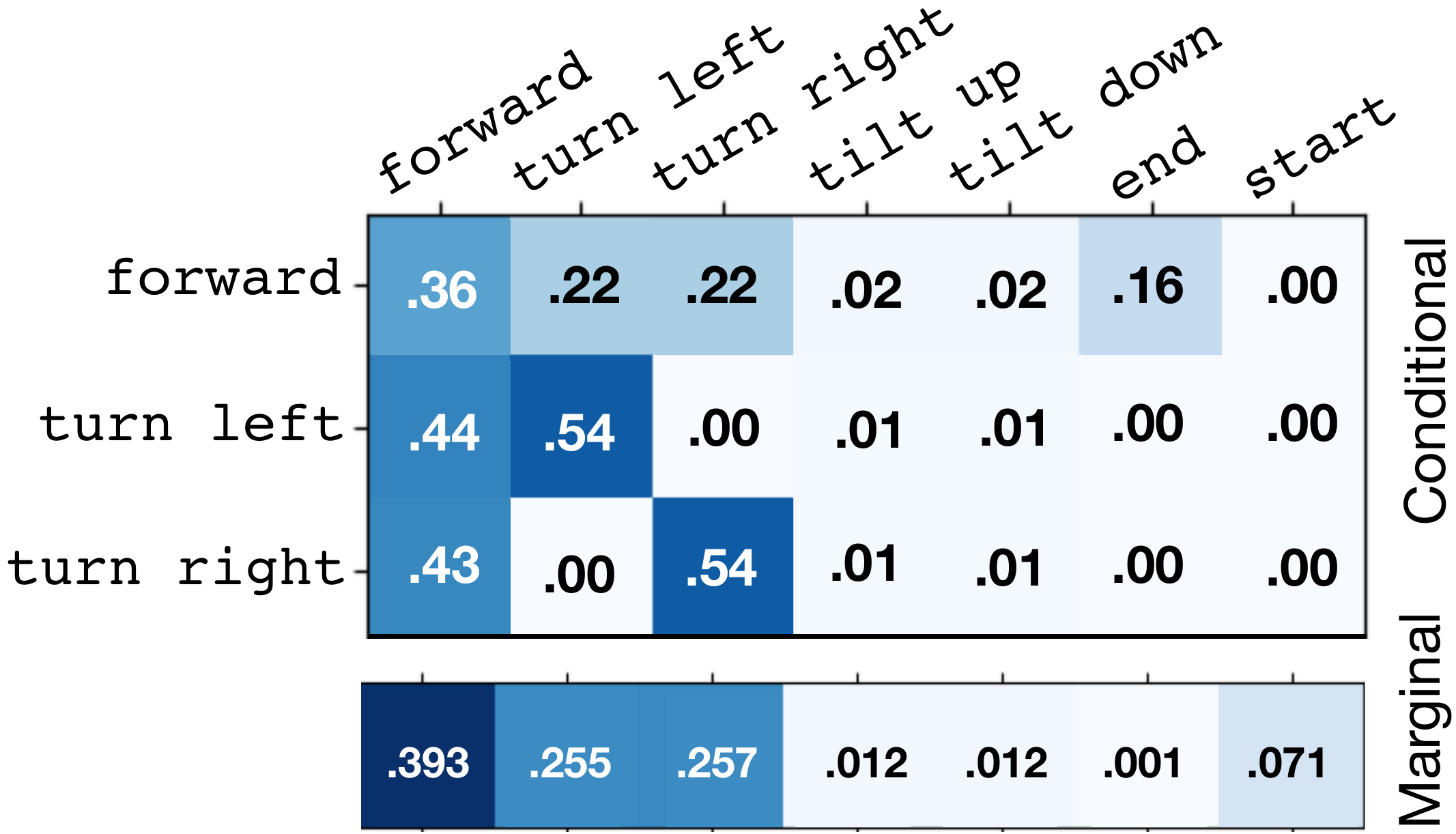}
\centering
\caption{$P(act=col | prev=row)$ and marginal action distributions in Matterport training.
Peaked distributions enable agents to memorize simple rules like not turning left immediately after turning right, or moving forward an average number of steps.
}
\label{fig:matterport_actions}
\end{figure}

\label{section:benchmarks}
We evaluate on navigation and question answering tasks across three benchmark datasets: Matterport Room-to-Room (no question answering component), and \iqadataset{} and EQA (question answering that requires navigating to the relevant scene in the environment)~\cite{anderson:cvpr18,gordon:cvpr18,das:eqa17}.
We divide the latter two into separate navigation and question answering components.
We then train and evaluate models separately per subtask to analyze accuracy.

\subsection{Matterport Room-to-Room}
An agent is given a route in English and navigates through a discretized map to the specified destination~\cite{anderson:cvpr18}.
This task includes high fidelity visual inputs and crowdsourced natural language routes.

\paragraph{Published Full Model:} At each timestep an LSTM decoder uses a ResNet-encoded image \visionin$_t$ and previous action \historyin$_{t-1}$ to attend over the states of an LSTM language encoder (\languagein) to predict navigation action \historyin$_t$ (seen in Figure~\ref{fig:matterport_actions}).

\paragraph{Published Baseline:} The agent chooses a random direction and takes up to five forward actions, turning right when no forward action is available.

\begin{figure}[t]
\includegraphics[width=.7\linewidth,trim={0 40pt 0pt 0},clip]{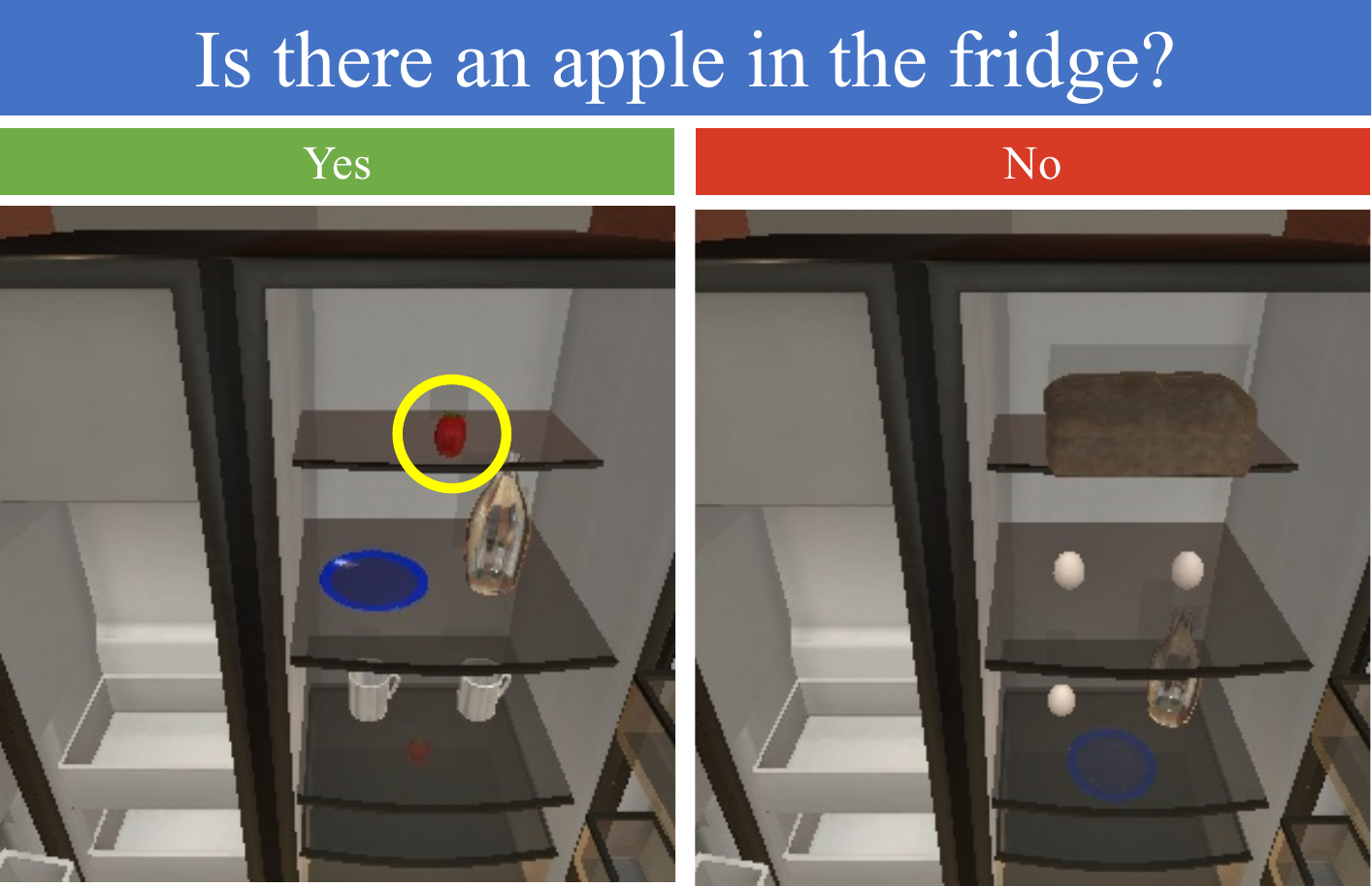}
\centering
\caption{IQA data construction attempts to make both the question and image necessary for QA.
}
\label{fig:iqa_example}
\end{figure}

\subsection{Interactive Question Answering}
\iqadataset{} \cite{gordon:cvpr18} contains three question types: existence (e.g., \emph{Is there a ...?}), counting (e.g., \emph{How many ...?}) where the answer ranges from 0 to 3, and spatial relation: (e.g., \emph{Is there a ... in the ...?}).
The data was constructed via randomly generated configurations to weaken majority class baselines (Figure~\ref{fig:iqa_example}).
To evaluate the navigation subtask, we introduce a new \thornav{} benchmark.\footnote{Formed from a subset of \iqadataset{} questions.}
The agent is placed in a random location in the room and must approach one of {\it fridge, garbage can,} or \textit{microwave} in response to a natural language question.

Although we use the same full model as \newcite{gordon:cvpr18}, our QA results are not directly comparable.
In particular, \newcite{gordon:cvpr18} do not quantify the effectiveness of the QA component independent of the scene exploration (i.e. navigation and interaction).
To remove the scene exploration steps of \newcite{gordon:cvpr18}, we provide a complete ground-truth view of the environment.\footnote{This approximates the agent having visited every possible location, interacted with all possible objects, and looked in all possible directions before answering.}
We use ground-truth rather than YOLO~\cite{redmon:cvpr16} due to speed constraints.

\paragraph{Nav Full Model:} The image and ground-truth semantic segmentation mask \visionin$_t$, tiled question \languagein, and previous action \historyin$_{t-1}$ are encoded via a CNN which outputs a distribution over actions. Optimal actions are learned via teacher forcing.

\paragraph{Nav Baseline:} The agent executes 100 randomly chosen navigation actions then terminates. In AI2THOR~\cite{ai2thor}, none of the kitchens span more than 5 meters. With a step-size of 0.25 meters, we observed that 100 actions was significantly shorter than the shortest path length.

\paragraph{Published QA Full Model:} The question encoding \languagein{} is tiled and concatenated with a top-down view of the ground truth location of all objects in the scene $\mvisionin$. This is fed into several convolutions, a spatial sum, and a final fully connected layer which outputs a likelihood for each answer. 

\paragraph{Published QA Baseline:} We include the majority class baseline from \newcite{gordon:cvpr18}.

\subsection{Embodied Question Answering}
EQA \cite{das:eqa17} questions are programmatically generated to refer to a single, unambiguous object for a specific environment, and are filtered to avoid easy questions (e.g., \emph{What room is the bathtub in?}).
At evaluation, an agent is placed a fixed number of actions away from the object.

\paragraph{Published Nav Full Model:} At each timestep, a planner LSTM takes in a CNN encoded image \visionin$_t$, 
LSTM encoded question \languagein, and the previous action \historyin$_{t-1}$ and emits an action \historyin$_t$. The action is executed in the environment, and then a lower-level controller LSTM continues to take in new vision observations and \historyin$_t$, either repeating \historyin$_t$ again or returning control to the planner.

\paragraph{Published Nav Baseline:} This baseline model is trained and evaluated with the same inputs as the full model, but does not pass control to a lower-level controller, instead predicting a new action using the planner LSTM at each timestep (i.e., no hierarchical control).
\newcite{das:eqa17} name this baseline \emph{LSTM+Question}.

\paragraph{Published QA Full Model:} Given the last five image encodings along the gold standard navigation trajectory, $\mvisionin_{t-4}\dots\mvisionin_{t}$, and the question encoding \languagein, image-question similarities are calculated via a dot product and converted via attention weights to a summary weight $\bar{\mvisionin}$, which is concatenated with \languagein{} and used to predict the answer.
\newcite{das:eqa17} name this oracle-navigation model \emph{ShortestPath+VQA}.

\paragraph{QA Baseline:} \newcite{das:eqa17} provide no explicit baseline for the VQA component alone.
We use a majority class baseline inspired by the data's entropy based filtering.

\section{Experiments}

\begin{table}
\centering
\begin{small}
\begin{tabular}{@{}l@{\hspace{5pt}}l@{\hspace{5pt}}r@{\hspace{5pt}}r@{\hspace{13pt}}r@{\hspace{5pt}}r@{\hspace{13pt}}r@{\hspace{0pt}}}
    & & \multicolumn{2}{@{}c@{\hspace{10pt}}}{\textbf{Matterport}$\uparrow$} & \multicolumn{2}{@{}c@{\hspace{10pt}}}{\bf \thornav{}$\uparrow$} & \multicolumn{1}{@{}c@{}}{\bf EQA$\downarrow$}\\
    & & \multicolumn{2}{@{}c@{\hspace{10pt}}}{(\textit{\%})} & \multicolumn{2}{@{}c@{\hspace{10pt}}}{ \textit{(\%)}} & \multicolumn{1}{@{}c@{}}{ \textit{(m)}}\\
    & \bf Model & \bf Seen & \textbf{Un} & \bf Seen & \textbf{Un} &  \textbf{Un}\\
    \toprule
	\multirow{2}{*}{\rotatebox[origin=c]{90}{Pub.}} & Full Model & 27.1 & 19.6 & 77.7 & 18.08 & 4.17 \\ 
	& Baseline & 15.9 & 16.3 & \phantom{0}2.18 & \phantom{0}1.54 & 4.21 \\
	\cmidrule{2-7}
	\multirow{3}{*}{\rotatebox[origin=c]{90}{Uni}} 
	& \navA & 18.5 & 17.1 & \phantom{0}4.53 & \phantom{0}2.88 & 4.53  \\ 
	& \navV & 21.2 & 16.6 & \bf \phantom{0}35.6  & \bf \phantom{0}7.50 & \bf $^*$4.11 \\
	& \navL & \textbf{23.0} & $^*$\textbf{22.1} & \phantom{0}4.03 & \phantom{0}3.46 & 4.64 \\
	\cmidrule[1pt]{1-7}
	$\Delta$ & Uni --  Base & \bad{\phantom{0}+7.1} & \bad{\phantom{0}+5.8} & \bad{+33.4} & \bad{\phantom{0}+5.96} & \bad{\phantom{0}-0.10} \\
	\bottomrule
\end{tabular}
\end{small}
\caption{
Navigation success (Matterport, \thornav{}) {\it (\%)} and distance to target (EQA) {\it (m)}.
{\bf Best} unimodal: \bad{better} than reported baseline; $^*$better than full model.
}
\label{tab:navigation}
\end{table}

Across all benchmarks, unimodal baselines outperform baseline models used in or derived from the original works.
Navigating unseen environments, these unimodal ablations outperform their corresponding full models on the Matterport (absolute $\uparrow2.5$\% success rate) and EQA ($\downarrow0.06m$ distance to target).

\subsection{Navigation}
\label{ssec:ex_nav}
We evaluate our ablation baselines on Matterport,\footnote{We report on Matterport-validation since this allows comparing Seen versus Unseen house performance.} \thornav{}, and EQA (Table~\ref{tab:navigation}),\footnote{For consistency with \thornav{} and EQA, we here evaluate Matterport using \emph{teacher forcing}.} and discover that some unimodal ablations outperform their corresponding full models.
For Matterport and \thornav{}, success rate is defined by proximity to the target.
For EQA, we measure absolute distance from the target in meters.

\paragraph{Unimodal Performance:}

Across Matterport, \thornav, and EQA, 
either \navV{} or \navL{} achieves better performance than existing baselines. 
In Matterport, the \navL{} ablation performs {\it better} than the Full Model in unseen environments. The diverse scenes in this simulator may render the vision signal more noisy than helpful in previously unseen environments.
The \navV{} model in \thornav{} and EQA is able to latch onto dataset biases in scene structure to navigate better than chance (for IQA), and the non-hierarchical baseline (in EQA).
In EQA, \navV{} also outperforms the Full Model;\footnote{EQA full \& baseline model performances do not exactly match those in \newcite{das:eqa17} because we use the expanded data updated by the authors \url{https://github.com/facebookresearch/EmbodiedQA/}.} the latent information about navigation from questions may be too distant for the model to infer.

The agent with access only to its action history (\navA) outperforms the baseline agent in Matterport and \thornav{} environments, suggesting it learns navigation correlations that are not captured by simple random actions (\thornav{}) or programmatic walks away from the starting position (Matterport).
Minimal sensing (which actions are available, $W$) coupled with the \emph{topological} biases in trajectories (Figure~\ref{fig:matterport_actions}), help this nearly zero-input agent outperform existing baselines.\footnote{This learned agent begins to relate to work in minimally sensing robotics \cite{OKane:2006}.}

\begin{table}
\centering
\begin{small}
\begin{tabular}{@{}l@{\hspace{5pt}}l@{\hspace{5pt}}r@{\hspace{5pt}}r@{\hspace{10pt}}}
    & & \multicolumn{2}{@{}c@{}}{{\bf Matterport}$\uparrow$} \\
    & & \multicolumn{2}{@{}c@{}}{(\textit{\%})} \\
    & \bf Model & \bf Seen & \bf Unseen \\
    \toprule
    \multirow{2}{*}{\rotatebox[origin=c]{90}{Pub.}} & Full Model & 38.6 & 21.8 \\
	& Baseline   & 15.9 & 16.3 \\
	\cmidrule{2-4}	
	\multirow{3}{*}{\rotatebox[origin=c]{90}{Uni}} & \navA & \phantom{0}4.1 & \phantom{0}3.2 \\ 
	& \navV & \textbf{30.6} & 13.3 \\
	& \navL & 15.4 & \bf 13.9 \\
	\cmidrule{2-4}
	$\Delta$ & Uni -- Base & \bad{+14.7} & \phantom{0}-2.4 \\
	\bottomrule
\end{tabular}
\end{small}
\caption{
Navigation results for Matterport when trained using student forcing.
{\bf Best} unimodal: \bad{better} than reported baseline.
}
\label{tab:navigation_sampling}
\end{table}

\paragraph{Matterport Teacher vs Student forcing}
With teacher forcing, at each timestep the navigation agent takes the gold-standard action regardless of what action it predicted, meaning it only sees steps along gold-standard trajectories.
This paradigm is used to train the navigation agent in \thornav{} and EQA.
Under student forcing, the agent samples the action to take from its predictions, and loss is computed at each time step against the action that would have put the agent on the shortest path to the goal.
Thus, the agent sees more of the scene, but can take more training iterations to learn to move to the goal.

Table~\ref{tab:navigation_sampling} gives the highest validation success rates across all epochs achieved in Matterport by models trained using student forcing.
The unimodal ablations show that the Full Model, possibly because with more exploration and more training episodes, is better able to align the vision and language signals, enabling generalization in unseen environments that fails with teacher forcing.

\begin{table}
\centering
\begin{small}
\begin{tabular}{@{}l@{\hspace{5pt}}l@{\hspace{10pt}}r@{\hspace{5pt}}r@{\hspace{5pt}}r@{\hspace{10pt}}r@{\hspace{5pt}}r@{\hspace{5pt}}r@{\hspace{5pt}}}
    & & \multicolumn{3}{@{}c@{}}{$\mathbf{d_T}\downarrow$} & \multicolumn{3}{@{}c@{}}{$\mathbf{d_{min}}\downarrow$} \\
    & & \multicolumn{3}{@{}c@{}}{(\textit{m})} & \multicolumn{3}{@{}c@{}}{(\textit{m})} \\
    & \bf Model & $T_{-10}$ & $T_{-30}$ & $T_{-50}$ & $T_{-10}$ & $T_{-30}$ & $T_{-50}$ \\
    \toprule
	\multirow{2}{*}{\rotatebox[origin=c]{90}{Pub.}} & Full & 0.971 & 4.17 & 8.83 & 0.291 & 2.43 & 6.45 \\
	& {\bf B}aseline & 1.020 & 4.21 & 8.73 & 0.293 & 2.45 & 6.38 \\ 
	\cmidrule{2-8}
	\multirow{3}{*}{\rotatebox[origin=c]{90}{{\bf U}ni}} & \navA & \bf $^*$0.893 & 4.53 & 9.56 & $^*$0.242 & 3.16 & 7.99 \\ 
	& \navV & $^*$0.951 & \bf $^*$4.11 & \bf $^{\dagger}$8.83 & $^*$0.287 & \bf 2.51 & \bf $^*$6.44 \\
	& \navL & 0.987 & 4.64 & 9.51 & \bf $^*$0.240 & 3.19 & 7.96 \\
	\cmidrule{2-8}
	$\Delta$ & U -- B & \bad{-0.127} & \bad{-0.10} & +0.10 & \bad{-0.053} & +0.06 & +0.06 \\
	\bottomrule
\end{tabular}
\end{small}
\caption{
Final distances to targets ($\mathbf{d_T}$) and minimum distance from target achieved along paths ($\mathbf{d_{min}}$) in EQA navigation.
{\bf Best} unimodal: \bad{better} than reported baseline; $^*$better than full model; $^{\dagger}$tied with full model.
}
\label{tab:all_nav_metrics}
\end{table}

\paragraph{EQA Navigation Variants}
Table~\ref{tab:all_nav_metrics} gives the average final distance from the target ($\mathbf{d_T}$, used as the metric in Table~\ref{tab:navigation}) and average minimum distance from target achieved along the path ($\mathbf{d_{min}}$) during EQA episodes for agents starting 10, 30, and 50 actions away from the target in the EQA navigation task.
At 10 actions away, the unimodal ablations tend to outperform the full model on both metrics, possibly due to the shorter length of the episodes (less data to train the joint parameters).
The \navV{} ablation performs best among the ablations, and ties with or outperforms the Full Model in all but one setting, suggesting that the EQA Full Model is not taking advantage of language information under any variant.

\subsection{Question Answering}
We evaluate our ablation baselines on \iqadataset{} and EQA, reporting top-1 QA accuracy (Table \ref{tab:qa_performance}) given gold standard navigation information as \visionin.
These decoupled QA models do not take in a previous action, so we do not consider \qaA{} ablations for this task.

\paragraph{Unimodal Performance:}
On \iqadataset{}, due to randomization in its construction, model ablations perform nearly at chance.\footnote{Majority class and chance for \iqadataset{} both achieve 50\%, 50\%, 25\% when conditioned on question type; our Baseline model achieves the average of these.}
The \qaV{} model with access to the locations of all scene objects only improves by 2\% over random guessing.

\begin{table}
    \centering
    \begin{small}
    \begin{tabular}{@{}l@{\hspace{5pt}}lll@{\hspace{10pt}}l@{\hspace{0pt}}}
        & & \multicolumn{2}{c}{\bf \iqadataset{}$\uparrow$} & \multicolumn{1}{c}{\bf EQA $\uparrow$} \\
        & \bf Model & \bf Unseen & \bf Seen & \bf Unseen \\
        \toprule
        \multirow{2}{*}{\rotatebox[origin=c]{90}{Pub.}} & Full Model & \phantom{+}88.3 & \phantom{+}89.3 & \phantom{+}64.0\\ 
         & Baseline & \phantom{+}41.7 &  \phantom{+}41.7 & \phantom{+}19.8 \\
        \cmidrule{2-5}
        \multirow{2}{*}{\rotatebox[origin=c]{90}{Uni}}
        & \qaV & \textbf{\phantom{+}43.5} & \textbf{\phantom{+}42.8} & \phantom{+}44.2 \\
        & \qaL & \phantom{+}41.7 &  \phantom{+}41.7 & \textbf{\phantom{+}48.8} \\
        \cmidrule[1pt]{1-5}
        $\Delta$ & Uni -- Base & \bad{\phantom{0}+1.8} & \bad{\phantom{0}+1.1} & \bad{+29.0} \\
        \bottomrule
    \end{tabular}
    \end{small}
    \caption{Top-1 QA accuracy.
    {\bf Best} unimodal: \bad{better} than reported baseline.
    }
    \label{tab:qa_performance}
\end{table}

\begin{figure}
\centering
\includegraphics[width=\linewidth]{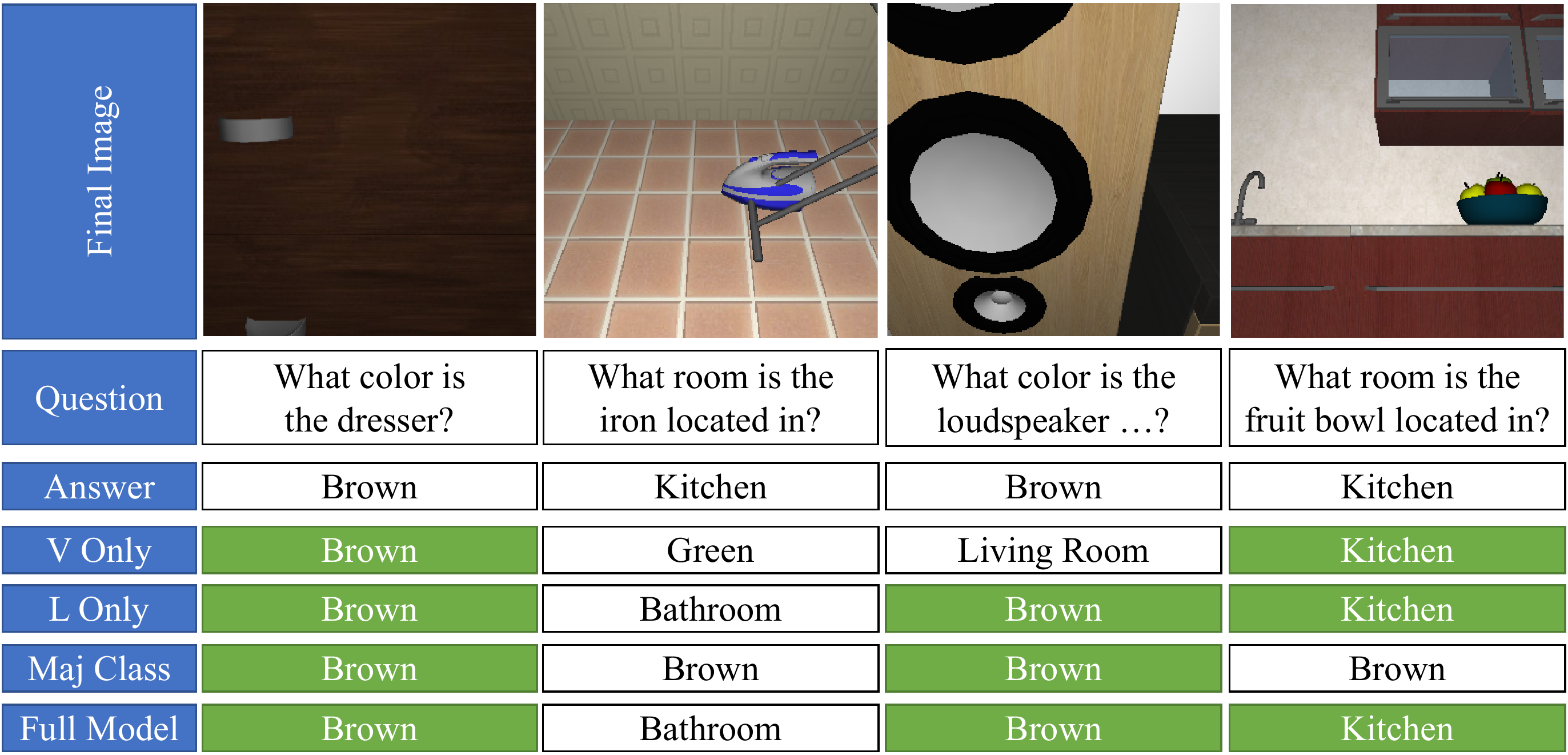}
\caption{Qualitative results on the EQA task. The language only model can pick out the most likely answer for a question.
The vision only model finds salient color and room features, but is unaware of the question.}
\label{fig:eqa_samples}
\end{figure}

For EQA, single modality models perform significantly better than the majority class baseline.
The vision-only model is able to identify salient colors and basic room features that allow it to reduce the likely set of answers to an unknown question.
The language only models achieve nearly 50\%, suggesting that despite the entropy filtering in \newcite{das:eqa17} each question has one answer that is as likely as all other answers combined (e.g. 50\% of the answers for \emph{What color is the bathtub?} are \emph{grey}, and other examples in Figure \ref{fig:eqa_samples}).

\section{Related Work}

Historically, semantic parsing was used to map natural language instructions to visual navigation in simulation environments~\cite{chen:aaai11,macmahon:aaai06}.
Modern approaches use neural architectures to map natural language to the (simulated) world and execute actions~\cite{paxton:icra19,chen:vigil18,nguyen:cvpr18,blukis:corl18,fried:nips18,mei:aaai16}.
In visual question answering (VQA)~\cite{antol:iccv15,Hudson:2019} and visual commonsense reasoning (VCR) \cite{zellers:2019}, input images are accompanied with natural language questions.
Given the question, egocentric QA requires an agent to navigate and interact with the world to gather the relevant information to answer the question. 
In both cases, end-to-end neural architectures make progress on these tasks. 

For language annotations, task design, difficulty, and annotator pay can introduce unintended artifacts which can be exploited by models to ``cheat'' on otherwise complex tasks~\cite{glockner-shwartz-goldberg:2018:Short, hypothesis-only-baselines-in-natural-language-inference}.
Such issues also occur in multimodal data like VQA~\cite{goyal18}, where models can answer correctly without looking at the image.
In image captioning, work has shown competitive models relying only on nearest-neighbor lookups~\cite{devlin2015exploring} as well as exposed misalignment between caption relevance and text-based metrics~\cite{rohrbach:emnlp18}.
Our unimodal ablations of visual navigation and QA benchmarks uncover similar biases, which deep architectures are quick to exploit.

\section{Conclusions}

In this work, we introduce an evaluation framework and perform the missing analysis from several new datasets.
While new state-of-the-art models are being introduced for several of these domains (e.g., Matterport: \cite{anonymous2019self-aware,ke:2019,Wang:2019,Ma:2019,tan:2019,fried:nips18}, and EQA: \cite{das:corl18}), they lack informative, individual unimodal ablations (i.e., ablating {\it both} language and vision) of the proposed models.

We find a performance gap between baselines used in or derived from the benchmarks examined in this paper and unimodal models, with unimodal models outperforming those baselines across all benchmarks.
These unimodal models can even outperform their multimodal counterparts.
In light of this, we recommend all future work include unimodal ablations of proposed multimodal models to vet and highlight their learned representations.

\section*{Acknowledgements}
This work was supported by NSF 
IIS-1524371, 1703166,  
NRI-1637479,
IIS-1338054, 1652052, 
ONR N00014-13-1-0720,
and the DARPA CwC program through ARO (W911NF-15-1-0543).   

\bibliography{main}
\bibliographystyle{acl_natbib}

\end{document}